\documentclass[conference]{IEEEtran}
\IEEEoverridecommandlockouts

\usepackage{cite}
\usepackage{amsmath,amssymb,amsfonts}
\usepackage{algorithmic}
\usepackage{graphicx}
\usepackage{textcomp}
\usepackage{xcolor}
\usepackage{booktabs}
\usepackage{multicol}
\usepackage{multirow}
\def\BibTeX{{\rm B\kern-.05em{\sc i\kern-.025em b}\kern-.08em
    T\kern-.1667em\lower.7ex\hbox{E}\kern-.125emX}}
\begin{document}

\title{Improving Micro-Expression Recognition with Phase-Aware Temporal Augmentation\\
}

\author{
    Vu Tram Anh Khuong,
    Luu Tu Nguyen, 
    Thanh Ha Le, 
    Thi Duyen Ngo\textsuperscript{*} \thanks{\textsuperscript{*}Corresponding author: duyennt@vnu.edu.vn} \\
    Faculty of Information Technology, VNU University of Engineering and Technology, Ha Noi, Viet Nam
}

\maketitle

\begin{abstract}
Micro-expressions (MEs) are brief, involuntary facial movements that reveal genuine emotions, typically lasting less than half a second. Recognizing these subtle expressions is critical for applications in psychology, security, and behavioral analysis. Although deep learning has enabled significant advances in micro-expression recognition (MER), its effectiveness is limited by the scarcity of annotated ME datasets. This data limitation not only hinders generalization but also restricts the diversity of motion patterns captured during training. Existing MER studies predominantly rely on simple spatial augmentations (e.g., flipping, rotation) and overlook temporal augmentation strategies that can better exploit motion characteristics. To address this gap, this paper proposes a phase-aware temporal augmentation method based on dynamic image. Rather than encoding the entire expression as a single onset-to-offset dynamic image (DI), our approach decomposes each expression sequence into two motion phases: onset-to-apex and apex-to-offset. A separate DI is generated for each phase, forming a Dual-phase DI augmentation strategy. These phase-specific representations enrich motion diversity and introduce complementary temporal cues that are crucial for recognizing subtle facial transitions. Extensive experiments on CASME-II and SAMM datasets using six deep architectures, including CNNs, Vision Transformer, and the lightweight LEARNet, demonstrate consistent performance improvements in recognition accuracy, unweighted F1-score, and unweighted average recall, which are crucial for addressing class imbalance in MER. When combined with spatial augmentations, our method achieves up to a 10\% relative improvement. The proposed augmentation is simple, model-agnostic, and effective in low-resource settings, offering a promising direction for robust and generalizable MER.

\end{abstract}


\section{Introduction}
\label{s: intro}
Micro-expressions (MEs) are brief, involuntary facial movements that reveal suppressed or concealed emotions. Typically lasting less than 0.5 seconds, these subtle expressions are difficult to detect both visually and algorithmically. Micro-expression recognition (MER) has garnered increasing interest due to its potential applications in psychology \cite{Bhushan2015}, security \cite{yan2013fast}, deception detection, and behavioral analysis \cite{polikovsky2010detection}. 

Although deep learning-based MER has attained impressive state-of-theart accuracy, surpassing human accuracy and other conventional approaches, the task remains highly challenging. The primary obstacles include the scarcity of annotated datasets, the subtle and transient nature of facial muscle activations, and the significant class imbalance within available data. Benchmark datasets such as CASME-II \cite{casme} and SAMM \cite{samm} together contain fewer than 400 sequences, making deep learning models prone to overfitting. Furthermore, commonly used spatial augmentations (e.g., flipping, rotation) increase visual diversity but do not enhance the temporal motion patterns that are critical for MER. To model temporal information, dynamic image (DI) has been used as a compact representation of motion through rank pooling. However, existing methods typically generate a single DI from the full sequence (onset-to-offset), which compresses both the rising (onset-to-apex) and falling (apex-to-offset) phases into a single representation. This conflation dilutes phase-specific motion cues and undermines temporal discrimination. 

This paper proposes a dual-phase dynamic image augmentation method that separately models the onset-to-apex and apex-to-offset segments. This phase-aware decomposition enriches the diversity of motion representations, preserves temporal progression, and enhances the model's ability to recognize subtle emotional transitions. In addition, our method is lightweight, model-agnostic, and easy to integrate into existing MER approaches. Extensive evaluations on the CASME-II and SAMM benchmarks across six deep architectures, including convolutional networks, Vision Transformers, and the lightweight LEARNet, demonstrate consistent improvements in accuracy, unweighted F1 (UF1) and unweighted average recall (UAR), confirming the effectiveness and generality of the proposed augmentation method.

The remainder of this paper is structured as follows. Section~\ref{s:RW} reviews related work on MER, data augmentation, and dynamic image representations. Next, section~\ref{s:method} details the proposed method. Then, section~\ref{s:experiment} presents experimental results and analyses. Finally, section~\ref{s:conclusion} concludes the paper and discusses future directions.

\section{Related Work}
\label{s:RW}

\subsection{Micro-Expression Recognition}
Traditional MER methods heavily relied on handcrafted features to capture subtle facial muscle movements. Techniques such as Local Binary Patterns from Three Orthogonal Planes (LBP-TOP) \cite{zhao2007dynamic}, Histogram of Oriented Gradients (HOG), optical flow \cite{polikovsky2010detection}, and optical strain \cite{liong2014optical} were commonly used. These descriptors, however, often required precise frame-level annotations (onset, apex, offset) and manual parameter tuning, limiting scalability. With the rise of deep learning, CNN-based approaches have dominated MER research. Many studies simplify MER by treating it as a static image classification task, using apex frames alone \cite{kim2016micro}. Although effective under data scarcity, this approach discards valuable temporal progression. Two-stream networks \cite{liu2016dual} and recurrent models such as LSTMs \cite{peng2017dual} have attempted to model temporal evolution, but require large amounts of data and are susceptible to overfitting.


\subsection{Motion Representations}
Optical flow has been a popular motion descriptor for MER, capturing pixel-wise displacements between frames. However, optical flow computation is sensitive to noise and small deformations. Furthermore, flow-based methods often require heavy computation and may suffer from error accumulation.

Dynamic images (DI) \cite{bilen2016dynamic} offer an alternative by summarizing motion through approximate rank pooling. Originally proposed for action recognition, DIs encode the temporal evolution of frames into a compact image, facilitating the use of 2D CNNs. In MER, dynamic images have been adopted as an alternative to optical flow for encoding subtle temporal motion while preserving spatial information. In particular, LEARNet~\cite{verma2019learnet} designed a lightweight CNN specifically for dynamic image input, achieving competitive results with fewer parameters. Similarly, works such as~\cite{le2020dynamic, verma2020non, zhou2023micro, tang2023novel} have utilized dynamic image techniques to summarize motion sequences, facilitating efficient MER model training. However, existing MER methods typically generate only a single dynamic image per sequence, spanning from onset to offset. This ignores the fact that MEs evolve through two distinct phases: onset-to-apex (expression rising) and apex-to-offset (expression fading). A single aggregation may overlook phase-specific motion patterns that are crucial for accurate interpretation. To address this, our work proposes a dual-phase dynamic image (dual-DI) strategy that better captures these expressive transitions.

\subsection{Data Augmentation for MER}
Data augmentation is critical for improving generalization in MER given the scarcity of annotated samples. Conventional techniques such as horizontal flipping, small-angle rotation, and cropping have been widely used \cite{liong2018less}. While effective for spatial regularization, these transformations do not introduce new temporal dynamics and may have limited impact on motion representation. More advanced augmentation strategies, including GAN-based synthetic data generation \cite{liong2020evaluation, yu2021ice}, frame interpolation \cite{li2019micro}, and adversarial training \cite{lee2022moma}, aim to enrich motion diversity. However, these approaches often require extensive tuning, are computationally expensive, and may introduce artifacts that compromise temporal consistency.


\begin{figure*}
    \centering
\includegraphics[width=0.8\linewidth]{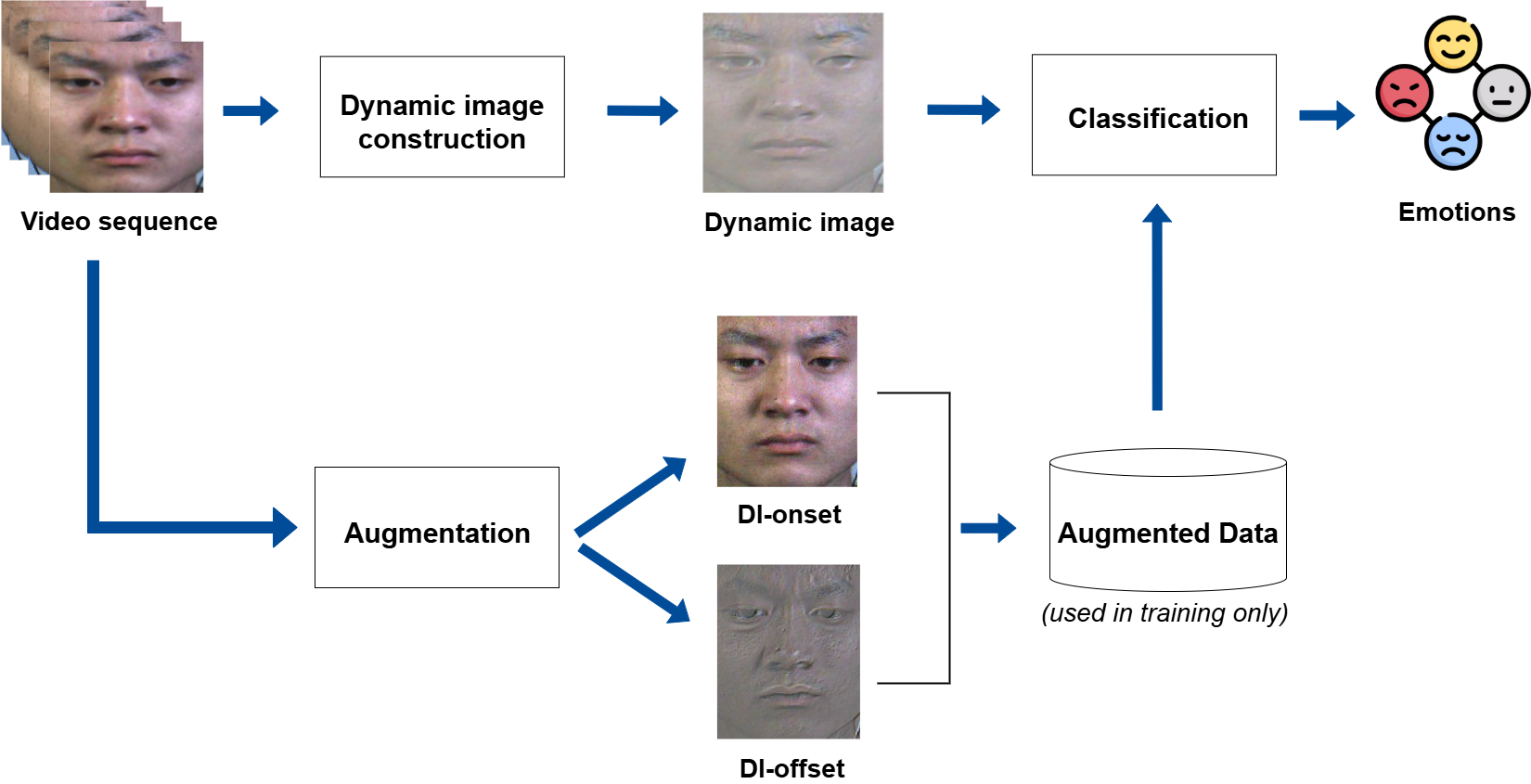}
    \caption{Overall pipeline of the proposed phase-aware dynamic image augmentation method}
    \label{fig: pipeline}
\end{figure*}

\section{Proposed Method}
\label{s:method}
The proposed MER approach enhances the effectiveness of dynamic images through an augmentation strategy that utilizes dual-phase dynamic imaging. By generating dynamic images for both the onset-to-apex and apex-to-offset segments, our method enriches the representational diversity of input data with temporally discriminative motion cues, as illustrated in Fig.~\ref{fig: pipeline}.

\subsection{Dynamic Image Construction}

\begin{figure}
    \centering
\includegraphics[width=0.7\linewidth]{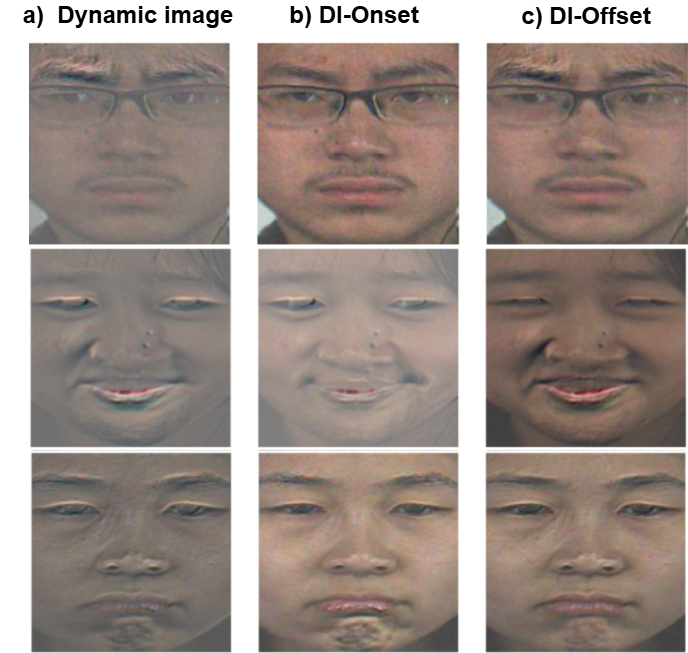}
    \caption{Visualization of dynamic image presentations (Best viewed in color).}
    \label{fig: DI}
\end{figure}


We adopt dynamic image construction via approximate rank pooling (ARP) \cite{bilen2016dynamic}, which summarizes temporal motion into a compact static image. Originally developed for action recognition, dynamic images have proven effective in summarizing complex motion patterns into compact, static representations. In the context of MER, this approach is particularly valuable, as MEs are inherently brief and subtle, making temporally aggregated representations more informative than analyzing individual frames.

Formally, given a sequence of $T$ consecutive frames ${F_1, F_2, \dots, F_T}$ and corresponding features $\psi(F_t)$, the dynamic image $d^*$ is computed as follows:

\begin{equation}
\label{eq:DI}
d^* = \sum_{t=1}^{T} \alpha_t \psi(F_t), \quad \text{with} \quad \alpha_t = 2t - T - 1.
\end{equation}

In our implementation, we use raw pixel values as features, i.e., $\psi(F_t) = F_t$.  The weighting coefficients $\alpha_t$ follow the formulation of ARP, assigning higher weights to later frames. This emphasizes temporal progression and encodes motion evolution into a compact static representation.

In this study, we apply Eq.\ref{eq:DI} to the onset-to-offset segment of each micro-expression, where $T$ denotes the number of frames between the annotated onset and offset. The resulting dynamic image, illustrated in Fig.\ref{fig: DI}a, summarizes the full temporal progression of the expression and is used as the primary input to the recognition model. To further improve temporal diversity and robustness, we generate additional phase-aware dynamic images derived from the onset-to-apex and apex-to-offset sub-segments. This augmentation strategy is detailed in Section~\ref{s: aug}.





\subsection{Augmentation}
\label{s: aug}


The standard ARP assigns increasing temporal weights to frames, implicitly assuming that motion evolves monotonically. While suitable for long action sequences, this assumption does not align with the dynamics of micro-expressions, which exhibit a brief, non-linear intensity pattern. Empirical studies and annotated datasets consistently show that micro-expressions reach peak intensity at the apex frame, followed by a rapid decline (see Fig.~\ref{fig: ME}).

To address this temporal asymmetry, we propose a dual-phase dynamic image augmentation strategy. Each micro-expression sequence is divided into two temporal segments (i.e., onset-to-apex and apex-to-offset) and Eq.~\ref{eq:DI} is applied separately to generate two distinct dynamic images:


\begin{figure}
    \centering
\includegraphics[width=0.8\linewidth]{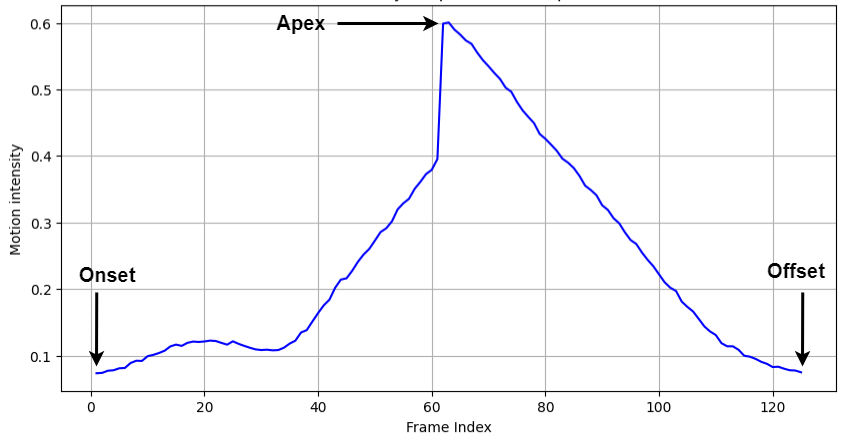}
    \caption{Micro-expression motion intensity graph}
    \label{fig: ME}
\end{figure}

\begin{itemize}
  \item \textbf{DI-Onset}: Encodes the onset-to-apex phase, where the facial motion gradually intensifies, as shown in Fig.~\ref{fig: DI}b. This dynamic image is computed using the standard ARP formulation, where frame weights increase linearly over time as defined in Eq.~\ref{eq:DI}. Here, $T$ denotes the number of frames in the onset-to-apex segment.
  
  \item \textbf{DI-Offset}: Encodes the apex-to-offset phase, corresponding to the relaxation of the expression, as shown in Fig.~\ref{fig: DI}c. To emphasize the apex frame and reflect the decrease in intensity, a reversed ARP is applied where frame weights decrease over time. The temporal weights for DI-Offset are defined as:
    \begin{equation} 
    \tilde{\alpha_t} = T + 1 - 2t. 
    \label{eq:alpha} 
    \end{equation}
where $T$ is the number of frames in the apex-to-offset segment, and $t$ indexes them sequentially. This weighting scheme mirrors the structure of standard ARP but inverses the temporal order, assigning the highest emphasis to the apex frame and gradually less to subsequent frames. As a result, DI-Offset captures the deceleration of facial motion while retaining the discriminative peak information.
\end{itemize}

This phase-aware augmentation enriches the training set with motion representations that reflect localized temporal structures without altering expression labels. Unlike conventional spatial transformations or synthetic generation methods, the proposed strategy introduces meaningful temporal variation at minimal computational cost. These augmented samples are used during training only and do not affect the inference pipeline.

\subsection{Classification}
The classification stage employs a diverse set of deep learning models to evaluate the generalization of DI across architectural paradigms. During training, each video yields three DI: the original onset-to-offset DI, and dual-phase DIs (DI-Onset and DI-Offset) generated through Dual-phase DI augmentation. Each of these dynamic images is treated as an independent training sample to enrich temporal diversity. However, only the original DI is used at evaluation.


The following backbones are used:
\begin{itemize}
\item VGG-Face~\cite{parkhi2015deep}: A VGG-16 variant pretrained on a large-scale face recognition dataset, known to transfer well to FER due to its rich facial representation.
\item VGG-19~\cite{simonyan2015very}: A classical CNN with 19 layers using stacked 3×3 convolution filters. Its depth and simplicity make it a strong baseline for static image-based FER.
\item ResNet-34~\cite{he2016deep}: Incorporates residual connections that allow gradients to flow through deeper layers, improving learning stability and expressiveness.
\item EfficientNet-B0~\cite{tan2019efficientnet}: A compact yet powerful model that scales width, depth, and resolution uniformly using compound coefficients.
\item ViT-B16~\cite{dosovitskiy2021image}: A vision transformer that replaces convolutions with multi-head self-attention, enabling global context modeling from fixed-size patches.
\item LEARNet~\cite{verma2019learnet}: A shallow but effective CNN designed specifically for MER using dynamic image inputs.
\end{itemize}

Except for LEARNet, all models were fine-tuned using transfer learning from pretrained weights, a common approach in MER to address the limited availability of labeled data. During training, models were optimized using the Adam optimizer with an initial learning rate of 0.0001, which was subsequently reduced using a cosine annealing schedule. Cross-entropy loss was consistently employed. LEARNet was implemented as described in the original paper \cite{verma2019learnet}.


\section{Experiments and Results}
\label{s:experiment}
A comprehensive set of experiments is conducted to validate the effectiveness of the proposed Dual-phase DI (dual-DI)  augmentation strategy. The objectives are threefold: (1) to assess the improvement in MER performance brought by Dual-phase DI augmentation compared to models trained without any augmentation, (2) to benchmark its effectiveness against conventional spatial augmentations such as horizontal flipping and rotation, and (3) to evaluate the generalization of the proposed approach across various deep learning architectures. All experiments are conducted under k-fold cross-validation protocols to ensure fairness, reproducibility, and alignment with established practices in MER research.
\subsection{Experiments}

\subsubsection{Datasets}

We evaluate our method on two widely used spontaneous micro-expression datasets: 
\begin{itemize}
    \item \textbf{CASME-II}~\cite{casme} includes 247 video sequences from 26 subjects, captured at 200 frames per second under controlled laboratory conditions. Each sample is manually annotated with onset, apex, and offset frames, along with one of five emotion categories: happiness, disgust, surprise, repression, and others
    \item \textbf{SAMM}~\cite{samm} consists of 159 sequences from 32 participants, also recorded at 200 fps but with high-resolution color imagery. Emotion labels were assigned by expert coders using the Facial Action Coding System (FACS), resulting in five classes: happiness, anger, surprise, contempt, and others. Each sequence includes precise temporal annotations (onset, apex, offset).
\end{itemize}

\subsubsection{Evaluation Protocol}

To evaluate model generalization across different subjects, k-fold cross-validation was employed \((k=5)\) - a widely used method in MER due to the typically limited availability of datasets \cite{review2}. The dataset is divided into \( k \) equal subsets, where each subset is used as the validation set once, while the remaining subsets are used for training. This process is repeated \( k \) times, and the final performance is computed as the mean and standard deviation of the metrics across all folds. This approach ensures robust and unbiased evaluation, maximizing the use of available data and providing a more reliable estimate of model performance.

\subsubsection{Evaluation Metrics}

The performance of each method is assessed using three standard metrics:
\begin{itemize}
\item \text{Accuracy}: Proportion of correctly classified samples.
\item \text{Unweighted F1-score (UF1)}: Macro-averaged F1 score, emphasizing balanced performance across classes.
\item \text{Unweighted Average Recall (UAR)}: Macro-averaged recall, accounting for per-class detection sensitivity.
\end{itemize}


\subsubsection{Implementation Details}
\label{s: imple}
\begin{table*}[t]
\centering
\caption{Performance on CASME-II and SAMM using different input configurations. Metrics reported as Accuracy (Acc), Unweighted F1 (UF1), and Unweighted Average Recall (UAR) in \%.}
\label{tab:results}
\begin{tabular}{l|l|ccc|ccc}
\toprule
\textbf{Model} & \textbf{Configuration} & \multicolumn{3}{c|}{\textbf{CASME-II}} & \multicolumn{3}{c}{\textbf{SAMM}} \\
 & & Acc & UF1 & UAR & Acc & UF1 & UAR \\
\midrule
\multirow{4}{*}{VGGFace} 
& Baseline & 61.22	&56.40&	54.69 & 55.56&	43.63	&43.07\\
& Baseline + Flip/Rotate & 64.90&2.97&	61.42 & 57.78	&42.46&	41.61 \\
& \textbf{Baseline + Dual-DI Aug} & \textbf{68.57}&65.59	&65.40 & 58.52	&44.44&	44.68\\
&  \textbf{Baseline + Dual-DI Aug + Flip/Rotate} & 67.35&	\textbf{66.11} & \textbf{65.81} & \textbf{59.26}&	\textbf{46.29}&	\textbf{45.94} \\
\midrule
\multirow{4}{*}{VGG-19} 
& Baseline & 55.92 & 49.14 & 47.12 & 50.74 & 32.75 & 35.92 \\
& Baseline + Flip/Rotate & 57.55 & 53.48 & 52.33 & 52.94 & 36.11 & 37.54 \\
& \textbf{Baseline + Dual-DI Aug} & 60.00 & 57.36 & 55.93 & 55.56 & 39.08 & 41.52 \\
& \textbf{Baseline + Dual-DI Aug + Flip/Rotate} & \textbf{60.41} & \textbf{57.67} & \textbf{58.10} & \textbf{57.04} & \textbf{41.78 }& \textbf{43.27} \\
\midrule
\multirow{4}{*}{ResNet-34} 
& Baseline & 55.10 & 50.73 & 53.29 & 48.15 & 35.17 & 35.57 \\
& Baseline + Flip/Rotate & 59.18 & 56.53 & 56.91 & 51.11 & 40.21 & 41.11 \\
&\textbf{Baseline + Dual-DI Aug} & 60.00 & 57.78 & 58.19 & 55.56 & 46.50 & 46.99 \\
& \textbf{Baseline + Dual-DI Aug + Flip/Rotate} &\textbf{62.45} & \textbf{58.78} & \textbf{59.76} & \textbf{56.30} & \textbf{45.15} & \textbf{45.03} \\
\midrule
\multirow{4}{*}{EfficientNet-B0} 
& Baseline & 53.06 & 48.44 & 48.95 & 48.89 & 36.44 & 37.26 \\
& Baseline + Flip/Rotate & 57.14 & 51.19 & 51.26 & 52.59 & 41.79 & 41.60 \\
& \textbf{Baseline + Dual-DI Aug} & 57.55 & 55.49 & 55.19 & 55.56 & 44.24 & 44.68 \\
&  \textbf{Baseline + Dual-DI Aug + Flip/Rotate} & \textbf{62.04} & \textbf{59.83} & \textbf{61.32} & \textbf{58.52}& \textbf{49.60} & \textbf{49.17} \\
\midrule
\multirow{4}{*}{ViT-B16} 
& Baseline & 54.69 & 50.27 & 50.11 & 55.56 & 43.57 & 44.18 \\
& Baseline + Flip/Rotate & 57.14 & 55.05 & 56.20 & 57.04 & 45.95 & 46.08 \\
& \textbf{Baseline + Dual-DI Aug} & 57.96 & 55.33 & 56.36 & \textbf{59.26} & \textbf{48.89} & \textbf{49.64 }\\
&  \textbf{Baseline + Dual-DI Aug + Flip/Rotate} & \textbf{61.63} & \textbf{59.82} & \textbf{59.98} & 57.78 & 46.56 & 47.47 \\
\midrule
\multirow{4}{*}{LEARNet*} 
& Baseline & 47.76 & 41.43 & 42.05 & 34.07 & 22.63 & 23.40 \\
& Baseline + Flip/Rotate & 48.57 & 44.55 & 44.50 & 38.15 & 26.52 & 27.91 \\
& \textbf{Baseline + Dual-DI Aug} & 50.61 & 46.26 & 46.06 & \textbf{47.41} & \textbf{36.84} & \textbf{37.25} \\
&  \textbf{Baseline + Dual-DI Aug + Flip/Rotate} & \textbf{56.33} & \textbf{53.94} & \textbf{53.57} & 41.85 & 31.40 & 31.41 \\
\bottomrule
\end{tabular}

\vspace{5pt} 
\small{ * This results are from our re-implementation with the model provided by the author \cite{verma2019learnet}}
\end{table*}

Dynamic images are resized to $224 \times 224$ pixels and normalized. Data augmentation strategies compared include:
\begin{itemize}
\item Baseline: Single dynamic image generated from the entire onset-to-offset sequence.
\item Baseline + Flip/Rotate: Horizontal flipping and random rotation within $\pm10^\circ$.
\item Baseline + Dual-DI Aug: Separate dynamic images generated for onset-to-apex and apex-to-offset phases.
\item Baseline + Dual-DI Aug + Flip/Rotate: Combination of temporal and spatial augmentations.
\end{itemize}

We evaluate six deep learning models: VGGFace, VGG-19, ResNet-34, EfficientNet-B0, ViT-B16, and LEARNet. Except for LEARNet, all models are initialized with ImageNet or face recognition pretrained weights and fine-tuned on the MER datasets. LEARNet is trained from scratch following its original design. Optimization is performed using the Adam optimizer with an initial learning rate of $10^{-4}$, reduced using cosine annealing. Batch size is set to 25, and models are trained for up to 50 epochs with early stopping based on validation loss to prevent overfitting.

\subsection{Results}

Table~\ref{tab:results} presents the recognition performance in terms of Accuracy, UF1, and UAR across six deep learning models on the CASME-II and SAMM datasets. Results are reported under four experimental configurations, as outlined in section~\ref{s: imple}.

The results demonstrate a consistent and substantial improvement when applying the proposed Dual-DI augmentation. Compared to the baseline without any augmentation, using Dual-DI alone increases average accuracy by 4.5\% on CASME-II and 6.5\% on SAMM. Corresponding improvements in unweighted F1-score (UF1) and unweighted average recall (UAR) are also observed: UF1 improves by 6.9\% and UAR by 6.3\% on CASME-II, while both metrics improve by 7.6\% on SAMM.

When compared specifically with the flip/rotate-only augmentation baseline, Dual-DI alone achieves higher accuracy by 3.7\% on CASME-II and 2.4\% on SAMM, with UF1 gains of 3.9\% and 3.2\%, and UAR gains of 3.5\% on both datasets. These results indicate that Dual-DI augmentation introduces complementary motion-aware information beyond what spatial transformations can provide.

Furthermore, combining Dual-DI with spatial augmentations leads to additional improvements on CASME-II, yielding average gains of 8.0\% in accuracy, 15.0\% in UF1, and 10.4\% in UAR. On the SAMM dataset, the combined strategy also improves performance in most cases, though a few exceptions are observed. In particular, for models like ViT-B16 and LEARNet, the addition of flipping and rotation slightly degrades performance compared to using Dual-DI alone. This suggests that spatial augmentations may disrupt the structured motion patterns captured by phase-aware dynamic images, especially in datasets with greater subject and recording variability like SAMM.

Among all configurations, the best overall result on CASME-II is achieved by VGGFace with Dual-DI augmentation and flip/rotate, reaching 67.35\% accuracy, 66.11\% UF1, and 65.81\% UAR. On SAMM, the highest performance is obtained with ViT-B16 using Dual-DI augmentation, achieving 59.26\% accuracy, 48.89\% UF1, and 49.64\% UAR. Notably, even VGGFace, a model pretrained for face recognition, benefits significantly from Dual-DI augmentation, outperforming all other models on CASME-II. This highlights the robustness of dynamic image representations across different model architectures. Finally, our LEARNet reimplementation, despite its shallow architecture, achieves notable gains: 8.57\% in accuracy and 12.51\% in UF1 on CASME-II when combined with Dual-DI augmentation and flip/rotate augmentation. This indicates that even lightweight models can benefit from motion-rich temporal augmentation.

These results prove the effectiveness of phase-aware dynamic image augmentation and demonstrate its potential as a general, model-agnostic enhancement for MER under limited training data scenarios.



\section{Conclusion}
\label{s:conclusion}
This paper has presented a phase-aware temporal augmentation strategy for MER. The proposed approach decomposes each expression sequence into onset-to-apex and apex-to-offset phases, generating dynamic images for each segment to capture the nuanced temporal progression inherent in micro-expressions. Such dual-phase augmentation enhances temporal diversity, provides richer motion representations, and complements traditional spatial augmentations. Experimental results across multiple deep learning architectures and two major MER datasets (CASME-II and SAMM) indicate that the proposed method consistently improves recognition performance. Specifically, the incorporation of dual-phase dynamic image augmentation leads to performance gains of up to 8\% in accuracy, 12\% in UF1, and 10\% in UAR compared to the baseline. These improvements are achieved when combining dual-phase augmentation with standard spatial augmentations, highlighting the substantial effectiveness of phase-aware modeling for MER. The method remains lightweight, model-agnostic, and easy to integrate into existing pipelines, making it particularly suitable for low-resource and real-world applications. Future work includes the exploration of adaptive phase segmentation, dynamic weighting strategies for enhanced temporal modeling, and the extension of this approach to other fine-grained facial analysis tasks beyond MER.

\bibliographystyle{ieeetr}
\bibliography{references}

\end{document}